\documentclass[letterpaper]{article} 
\usepackage{aaai23}  
\usepackage{times}  
\usepackage{helvet}  
\usepackage{courier}  
\usepackage[hyphens]{url}  
\usepackage{graphicx} 
\usepackage{natbib}  
\usepackage{caption} 
\usepackage{algorithm}
\usepackage{algorithmic}

\usepackage{newfloat}
\usepackage{listings}

\usepackage{multirow}
\usepackage{diagbox}
\usepackage{booktabs}
\usepackage{subfig}
\usepackage{bbding}
\usepackage{float}
\usepackage[inline]{enumitem}
\usepackage{amsmath}
\usepackage{makecell}

\DeclareCaptionStyle{ruled}{labelfont=normalfont,labelsep=colon,strut=off} 
\floatstyle{ruled}
\newfloat{listing}{tb}{lst}{}
\floatname{listing}{Listing}
\title{Detecting Multivariate Time Series Anomalies with Zero Known Label}

\author {
    Qihang	Zhou\textsuperscript{\rm 1},
    Jiming	Chen\textsuperscript{\rm 1},
    Haoyu	Liu\textsuperscript{\rm 1,\rm 2}\thanks{Corresponding author.},
    Shibo	He\textsuperscript{\rm 1,\rm 3},
    Wenchao	Meng\textsuperscript{\rm 1}
}

\affiliations {
    \textsuperscript{\rm 1}Zhejiang University, Hangzhou, China,\\
    \textsuperscript{\rm 2}NetEase Fuxi AI Lab, Hangzhou, China,\\
    \textsuperscript{\rm 3}Key Laboratory of Collaborative Sensing and Autonomous Unmanned Systems of Zhejiang Province, Hangzhou, China\\
    \{zqhang, cjm, haoyu\_liu, s18he, wmengzju\}@zju.edu.cn, liuhaoyu03@corp.netease.com
}

\begin{document}

\maketitle

\begin{abstract}
Multivariate time series anomaly detection has been extensively studied under the one-class classification setting, where a training dataset with all normal instances is required. However, preparing such a dataset is very laborious since each single data instance should be fully guaranteed to be normal. It is, therefore, desired to explore multivariate time series anomaly detection methods based on the dataset without any label knowledge.
In this paper, we propose MTGFlow, an unsupervised anomaly detection approach for \underline{M}ultivariate \underline{T}ime series anomaly detection via dynamic \underline{G}raph and entity-aware normalizing \underline{Flow}, leaning only on a widely accepted hypothesis that abnormal instances exhibit sparse densities than the normal.
However, the complex interdependencies among entities and the diverse inherent characteristics of each entity pose significant challenges to density estimation, let alone to detect anomalies based on the estimated possibility distribution. 
To tackle these problems, 
we propose to learn the mutual and dynamic relations among entities via a graph structure learning model, which helps to model the accurate distribution of multivariate time series. Moreover, taking account of distinct characteristics of the individual entities, an entity-aware normalizing flow is developed to describe each entity into a parameterized normal distribution, thereby producing fine-grained density estimation. Incorporating these two strategies, MTGFlow achieves superior anomaly detection performance. 
Experiments on five public datasets with seven baselines are conducted, MTGFlow outperforms the
SOTA methods by up to 5.0 AUROC\%.


\end{abstract}
\section{Introduction}
Multivariate time series (MTS) broadly exist in many important scenarios, such as production data produced by multiple devices in smart factories and monitoring data generated by various sensors in smart grids. Anomalies in MTS exhibit unusual data behaviors at a specific time step or during a time period.
To identify these anomalies, previous methods mostly focus on training one-class classification (OCC) models from only normal data~\cite{wu2021current, OC-SVM,su2019robust,chen2021daemon, deng2021graph,xu2021anomaly, zhang2019deep}.
They heavily rely on an assumption that the training dataset with all normal samples can be easily obtained~\cite{ruff2021unifying}.


However, this assumption may not always hold in real-world scenarios~\cite{goodge2021lunar, zhang2019online, zong2018deep, qiu2022latent}, leading to noisy training datasets with the mixture of normal and abnormal data instances. Meanwhile, it is already verified that model training procedure is prone to overfitting noisy labels~\cite{zhang2021understanding}, so that the performance of those OCC-based methods could be severely degraded~\cite{wang2019effective, huyan2021unsupervised}. Therefore, it is rewarding to develop unsupervised MTS anomaly detection methods based on the dataset with absolute zero known labels. 


An effective unsupervised strategy is modeling the dataset into a distribution, relying only on a widely accepted hypothesis that abnormal instances exhibit sparse densities than the normal, i.e., the low-density regions consist of abnormal samples and the high-density regions are formed by the normal samples~\cite{gupta2013outlier, Guansong, Ruoying}. Methods have been explored along side this strategy and the key challenge lies in the accurate density estimation of the distribution. 
Time series density is modeled as the parameterized probability distribution~\cite{salinas2020deepar, rasul2021autoregressive, feng2022multi},
while it is still challenging to model a more complex data distribution. To improve the model capacity of density estimation, Rasul~\emph{et al.}~\cite{rasul2020multivariate} further exploits normalizing flow to model complex distribution for high-dimensional MTS~\cite{rasul2020multivariate}. However, they neglect the interdependencies among constituent series which also play an important role in accurate density estimation.

The most related work is GANF~\cite{dai2021graph}, which tackles the same MTS anomaly detection task. In their design, the static directed acyclic graph (DAG) is leveraged to model intractable dependence among multiple entities, and normalizing flow~\cite{dinh2016density, papamakarios2017masked} is employed to estimate an overall distribution for all entities together. Although GANF has achieved state-of-the-art (SOTA) results previously, it still suffers from two drawbacks. 
First, rather than a static inter-relationship, in real-world applications, the mutual dependencies among entities could not only be complex but also evolving. This dynamic property can not be simply characterized via a DAG structure. Second, entities usually have diverse working mechanisms, leading to diverse sparse characteristics when anomalies occurred. GANF projects all entities into the same distribution, resulting in a compromise for the density estimation of each individual time series. Thereby, the final anomaly detection performance could also be degraded.



In this paper, we propose MTGFlow, an unsupervised
anomaly detection method for MTS anomaly detection, to tackle the above problems. First, considering the evolving relations among entities, we introduce a graph structure learning module to model these changeable interdependencies. To learn the dynamic structure, a self-attention module~\cite{vaswani2017attention} is plugged into our model for its superior performance on quantifying pairwise interaction. 
Second, aiming at the diverse inherent characteristics existed among individual entities, we design an entity-aware normalizing flow to model the entity-specific density estimation. Thereby, each entity can be assigned to a unique target distribution and the diverse entity densities can be estimated independently.
In addition, we also propose to control the model size by sharing entity-specific model parameters, which helps MTGFlow achieve fine-grained density estimation without much memory consumption.

We summarize our contributions as follows:

\begin{itemize}

\item We propose MTGFlow, a new SOTA method for unsupervised MTS anomaly detection. It essentially enables anomaly localization and interpretation.


\item We model the complicated interdependencies among entities into the dynamic graph, which captures the complex and evolving mutual dependencies among entities. Also, entity-aware normalizing flow is introduced to produce entity-specific density estimation.

\item Experiments on five datasets with seven baseline methods are conducted, outperforming the
SOTA methods by up to 5.0 AUROC\%. 




\end{itemize}

\section{Related Work}
\subsection{Time Series Anomaly Detection}
Time series anomaly detection has been extensively investigated under OCC setting~\cite{chalapathy2019deep,hundman2018detecting}. Previous influential methods like DeepSVDD~\cite{ruff2018deep}, EncDecAD~\cite{malhotra2016lstm},
OmniAnomaly~\cite{su2019robust}, USAD~\cite{audibert2020usad} and DAEMON~\cite{chen2021daemon} firstly train a model with absolutely normal instances so that the abnormal instances would exhibit differently when they are fed into the model during testing. Along side this line, Anomaly Transformer~\cite{xu2021anomaly} and TranAD~\cite{tuli2022tranad} further investigate fine-grained representation learning procedure via techniques like self-attention~\cite{vaswani2017attention}, achieving good detection performance.


However, all these works are based on the assumption that a sufficient training dataset with all normal instances can be acquired, which is very hard for real-world applications since each instance should be manually checked carefully. In addition, once there exist abnormal instances in the training data, the performance of these OCC-based detection methods could be severely degraded~\cite{wang2019effective, huyan2021unsupervised}. Therefore, instead of fitting distribution of the normal training dataset, Dai and Chen propose GANF~\cite{dai2021graph} to detect MTS anomalies in an unsupervised manner. Inspired by them, we propose MTGFlow to facilitate the learning capacity and improve detection performance.

\subsection{Graph Structure Learning}
Graph convolution networks (GCN)~\cite{kipf2016semigcn} have achieved great success in modeling intrinsic structure patterns. However, such a graph structure is usually unknown in real-world scenarios so graph structure learning methods are in need~\cite{velivckovic2017graphgat}. In the field of anomaly detection of MTS, some recent works attempt to explore this area. GDN~\cite{deng2021graph} learns a directed graph via node embedding vectors. According to the cosine similarity of embedding vectors, top-K candidates of each node are considered to have interdependencies. GANF~\cite{dai2021graph} models relations among multiple sensors, using DAG, and learns the structure of the DAG through continuous optimization with a simplified constraint that facilitates backward propagation. Our work MTGFlow models the mutual complex dependence as a fully connected graph via the self-attention mechanism, so that a much more flexible relation among entities can be represented.


\subsection{Normalizing Flow for Anomaly Detection}

Normalizing flow is an important technology on density estimation and has been successfully utilized in image generation tasks~\cite {dinh2016density,papamakarios2017masked}. Recently, it is also explored for anomaly detection based on the assumption that anomalies are in low-density regions, like DifferNet~\cite{rudolph2021same} and CFLOW-AD~\cite{gudovskiy2022cflow}, which leverage normalizing flow to estimate likelihoods of normal embeddings and declare image defects when the embedding is far away from the dense region. GANF is a great work that employs normalizing flow for unsupervised MTS anomaly detection. We follow this research line and facilitate model capacity through an entity-aware normalizing flow design.

\begin{figure*}
    \centering
    \includegraphics[width=1\textwidth]{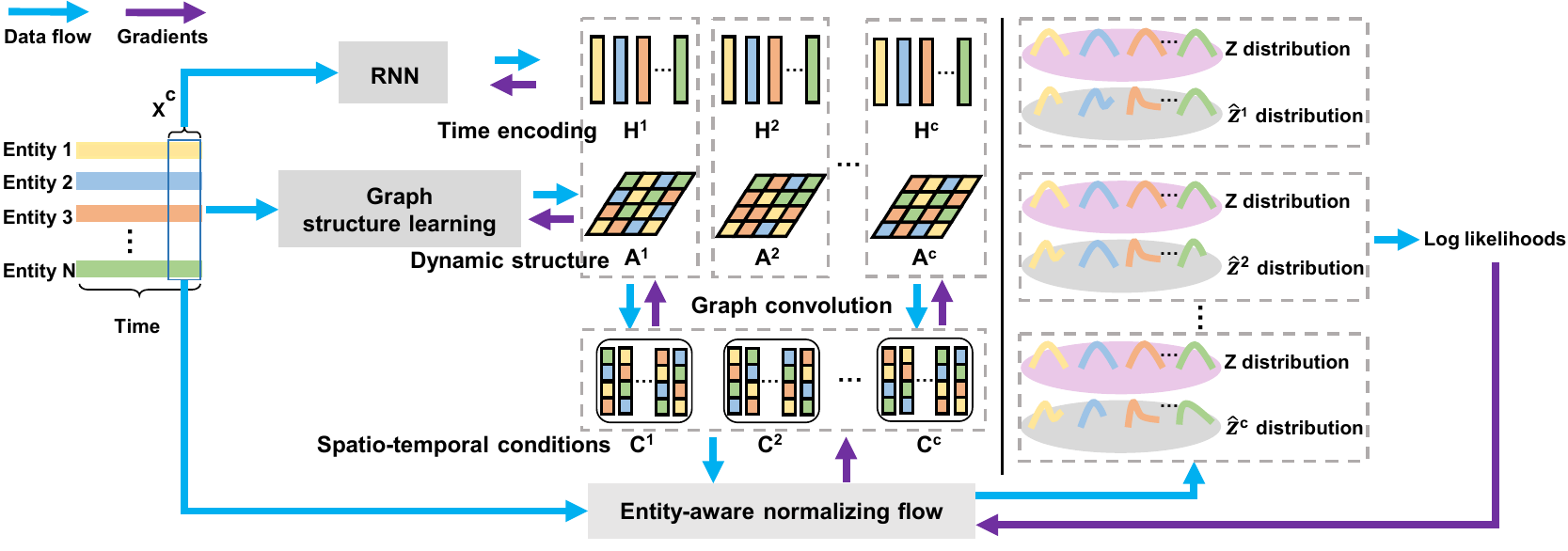}
    \caption{Overview of the proposed MTGFlow. 
    Within a sliding window of size $T$, time series $x^{c}$ is fed to the RNN module to capture the temporal correlations.
    Hidden states of RNN are regarded as time encoding, $H^{c}$. Meanwhile, $x^{c}$ is also input to the graph structure learning module to capture dynamic interdependencies among entities, which are modeled as adjacency matrix $A^{c}$. The spatio-temporal conditions $C^{c}$ are derived via the graph convolution operation for $H^{c}$ and $A^{c}$. Finally, $C^{c}$ is used to help entity-aware normalizing flow model to produce entity-specific density estimation for the distribution of time series.}
    \label{fig:overview}
\end{figure*}
\section{Preliminary}
In this section, we give a brief introduction of normalizing flow to better understand MTGFlow.


\subsection{Normalizing Flow}
Normalizing flow is an unsupervised density estimation approach to map the original distribution to an arbitrary target distribution by the stack of invertible affine transformations. 
When density estimation on original data distribution $\mathcal{X}$ is intractable, an alternative option is to estimate $z$ density on target distribution $\mathcal{Z}$. Specifically, suppose a source sample $x \in \mathcal{R}^{D} \sim \mathcal{X}$ and a target distribution sample $z \in \mathcal{R}^{D} \sim \mathcal{Z}$. Bijective invertible transformation $\mathcal{F}_{\theta}$ aims to achieve one-to-one mapping $z = f_{\theta}(x)$ from $\mathcal{X}$ to $\mathcal{Z}$. According to the change of variable formula, we can get 
$
     P_\mathcal{X}(x) = P_\mathcal{Z}(z)\left |\det\frac{\partial{f_{\theta}}}{\partial{x}^{T}}\right|.
$
    Benefiting from the invertibility of mapping functions and tractable jacobian determinants $\left |\det\frac{\partial{f_{\theta}}}{\partial{x}^{T}}\right|$.
    The objective of flow models is to achieve $\hat{z} = z$, where $\hat{z} = f_{\theta}(x)$.
Flow models are able to achieve more superior density estimation performance when additional conditions $C$ are input~\cite{ardizzone2019guided}. 
Such a flow model is called conditional normalizing flow, and its corresponding mapping is derived as $z = f_{\theta}(x|C)$. Parameters $\theta$ of ${f}_{\theta}$ are updated by maximum likelihood estimation (MLE): $\theta^{*}=\mathop{\arg\max}\limits_{\theta}(log(P_\mathcal{Z}(f_{\theta}(x|C)) + log(\left |\det\frac{\partial{f_{\theta}}}{\partial{x}^{T}} \right|))$




\section{Method}

\subsection{Data Preparation}
MTS are defined as $x=\left( x_1, x_2, ..., x_K \right)$ and $x_i \in \mathcal{R}^{L}$, where $K$ represents the total number of entities, and $L$ denotes the total number of observations of each entity. We use the z-score to normalize the time series from different entities.
$
     \bar x_i = \frac{x_i - mean(x_i)}{std(x_i)}
$,
where $mean(x_i)$ and $std(x_i)$ represent the mean and standard deviation of the i-$th$ entity along the time dimension, respectively. To preserve temporal correlations of the original series, we use a sliding window with size $T$ and stride size $S$ to sample the normalized MTS. $T$ and $S$ can be adjusted to obtain the training sample $x^{c}$, where c is the sampling count. $x^{c}$ is short for $x^{cS:cS+T}$.


\subsection{Overall Structure}
The core idea behind MTGFlow is to dynamically model mutual dependence so that fine-grained density estimation of the multivariate time series can be obtained.
Such accurate estimation enables the superiority of capturing low-density regions, and further promotes the anomaly detection performance even if there is high anomaly contamination in the training dataset.
Fig. \ref{fig:overview} shows the overview of MTGFlow. In particular, we model the temporal variations of each entity, using RNN model. Meanwhile, a graph structure learning module is leveraged to model the dynamic interdependencies. Then, the derived time encoding, output of RNN, performs the graph convolution operation with the above corresponding learned graph structure. We regard above outputs as spatio-temporal \emph{conditions} as they contain temporal and structural information. Next, the spatio-temporal conditions are input to help entity-aware normalizing flow achieve precise fine-grained density estimation. The deviations of $\hat{z}$ and $z$ are measured by log likelihoods. Finally, all modules of MTGFlow are jointly optimized through MLE. 

\subsection{Graph Structure Learning via Self-attention}

Since dependence among entities is mutual and evolves over time, we exploit self-attention to learn a dynamic graph structure. Entities in multivariate time series are regarded as graph nodes. Given the window sequence $x^{c}$, the query and key of node $i$ are represented by vectors $x_{i}^{c}W^Q$ and $x_{i}^{c}W^K$, where $W^Q\in \mathcal{R}^{T \times T}$ and $W^K\in \mathcal{R}^{T \times T}$ are the query and key weights. The pairwise relationship $e_{ij}^{c}$ at the c-$th$ sampling count between node i and node j is described as $e_{ij}^{c} = \frac{(x_{i}^{c}W^Q)(x_{j}^{c}W^K)^{T}}{\sqrt{T}}$. The attention score $a_{ij}^{c}$ is used to 
quantify the pairwise relation from node $i$ to node $j$, calculated by $a_{ij}^{c} = \frac{\exp (e_{ij}^{c})}{\sum_{j = 1}^{K}exp({e_{ij}^{c}})}$.
And the attention matrix consists of attention scores of each node, thus including mutual dependence among entities. Naturally, we treat the attention matrix as the adjacency matrix $A^{c}$ of the learned graph.
Since input time series are evolving over time, $A^{c}$ also changes to capture the dynamic interdependencies.

\subsection{Spatio-temporal Condition}
To better estimate the density of multiple time series, the robust spatio-temporal condition information is important. As described above, underlying structure information is modeled as the dynamic graph. Besides spatio information, temporal correlations also play an important role to feature time series. 
Here, we follow the most prevalent idea, where RNN is utilized to capture the time correlations.
For a window sequence of entity $k$, $x^{c}$, the time representation $H_{k}^{t}$ at time $t\in [cS:cS+T)$ is derived by $H_{k}^{t} = \textbf{RNN}({x_{k}^{t}, H_{k}^{t-1}})$, where RNN can be any sequence model such as LSTM~\cite{hochreiter1997long} and GRU~\cite{cho2014learning}, and $H_{k}^{t}$ is the hidden state of RNN. 
To derive the spatio and temporal information $C^{t}$ of all entities at $t$, a graph convolution operation is performed through the learned graph $A^{c}$. As mentioned in GANF, we also find that history information of the node itself helps enhance temporal relationships of time series. Hence, the spatio-temporal condition at $t$:
\begin{small}
\begin{equation}
    C^{t} = ReLU(A^{c}H^{t}W_{1} + H^{t-1}W_{2})W_{3},
\end{equation}
\end{small}
where $W_{1}$ and $W_{2}$ are graph convolution and history information weights, respectively. $W_{3}$ is used to improve the expression ability of condition representations. The spatio-temporal condition $C^{c}$ for window $c$ is the concatenation of $C^{t}$ along the time axis.
\subsection{Entity-aware Normalizing Flow}

Distributions of individual entities have discrepancies because of their different work mechanisms, and thus their respective anomalies will generate distinct sparse characteristics.
 If we map time series from all entities to the same distribution $N(0,I)$, as does in GANF, then the description capacity of the model will be largely limited and the unique inherent property of each entity will be ignored. 
Therefore, we design the entity-aware normalizing flow $z_k = f^{k}_{\theta}(x|C)$ to make more detailed density estimation, where $x$, $C$, $k$ are the input sequence, condition, and the k-$th$ entity, respectively. 
Technically, 
for one entity, we assign the multivariate Gaussian distribution as the target distribution. The covariance matrix of the above target distribution is the identity matrix $I$ for better convergence. Moreover, in order to generate different target distributions $\mathcal{Z}_k$, we independently draw mean vectors $\mu_k \in R^{T}$ from $N(0,I)$~\cite{izmailov2020semi}, However, we find that such setting results in performance degradation. So, in our experiment, each element of $\mu_k$ is kept the same. Specifically, for the time series of the entity k, the density estimation is given by:
\begin{small}
\begin{equation}
    \begin{split}
    P_{\mathcal{X}_k}(x_k) = P_{\mathcal{Z}_k}&(f_{\theta}^{k}(x_k|C))\left       |\det\frac{\partial{f_{\theta}^{k}}}{\partial{x_k}^{T}} \right|\\
    \mathcal{Z}_k & =  N(\mu_k, I) \\
    \end{split}
\end{equation}
\end{small}
where each element of $\mu_k$ is the same, and is drawn from the $N(0, 1)$. In such a case, model parameters will increase with the number of entities. To mitigate this problem, we share entity-aware normalizing flow parameters across all entities. So, the density estimation for k reads:
\begin{small}
\begin{equation}
    P_{\mathcal{X}_k}(x_k) = P_{\mathcal{Z}_k}(f_{\theta}(x_k|C))\left       |\det\frac{\partial{f_{\theta}}}{\partial{x_k}^{T}} \right|
\end{equation}
\end{small}

\subsection{Joint Optimization}
As described above, MTGFlow combines graph structure learning and RNN to capture the spatio and temporal dependence on multiple time series. Then, derived saptio-temporal conditions are utilized to contribute to entity-aware normalizing flow accurately estimating density of time series. To avoid getting stuck in the local optimum for each module of MTGFlow, we jointly optimize all modules for overall performance of MTGFlow. The whole parameters $W^*$ are estimated via MLE.
\begin{small}
\begin{equation*}
    \begin{split}
        &W^*=\mathop{\arg\max}\limits_{W}log(P_\mathcal{X}(x)) \\
        &\approx\mathop{\arg\max}\limits_{W}\frac{1}{NK}\sum^N_{c=1}\sum^K_{k=1}log(P_{\mathcal{Z}_k}(f_{\theta}(x_{k}^{c}|C_{k}^{c}))\left|\det\frac{\partial{f_{\theta}}}{\partial{x_{k}^{c}}^{T}} \right|)\\
        &\approx \mathop{\arg\max}\limits_{W}\frac{1}{NK}\sum^N_{c=1}\sum^K_{k=1}-\frac{1}{2}\lVert \hat{z_{k}^{c}} -\mu_k\rVert_{2}^2 + log\left|\det\frac{\partial{f_{\theta}}}{\partial{x_{k}^{c}}^{T}} \right|,
    \end{split}
\end{equation*}
\end{small}
where $N$ is the total number of windows.

\subsection{Anomaly Detection and Interpretation}
Based on the hypothesis that anomalies tend to be sparse on data distributions, low log likelihoods indicate that the observations are more likely to be anomalous. 
\subsubsection{Anomaly Detection}
Taking the window sequence $x_{k}^{c}$ as the input, the density of all entities can be estimated.
The mean of the negative log likelihoods of all entities serves as the anomaly score $S_c$, which is calculated by:
\begin{small}
\begin{equation}
    S_c =  -\frac{1}{K}\sum^K_{k=1}log(P_{\mathcal{X}_k}(x_{k}^{c}))
    \label{equ:anomalyscore}
\end{equation}
\end{small}
A higher anomaly score represents that $x_{k}^{c}$ locates in the lower density region, indicating a higher possibility to be abnormal. Since abnormal series exist in the training set and validation set, we cannot directly set the threshold to label the anomaly, such as the maximum deviation in validation data~\cite{deng2021graph}. Therefore, to reduce the anomaly disturbance, we store $S_c$ of the whole training set, and the interquartile range (IQR)  is used to set the threshold: $
    Threshold = Q_3 + 1.5*(Q_3 - Q_1),
$
where $Q_1$ and $Q_3$ are 25-$th$ and 75-$th$ percentile of $S_{c}$. 
\subsubsection{Anomaly Interpretation}
Abnormal behaviors of any entity could lead to the overall abnormal behavior of the whole window sequence.
Naturally, we can get the entity anomaly score $S_{ck}$ for entity k according to Eq.~\eqref{equ:anomalyscore}.
\begin{small}
\begin{equation}
    S_c =  -\frac{1}{k}\sum^K_{k=1}log(P_{\mathcal{X}_k}(x_{k}^{c})) = \sum^K_{k=1}S_{ck}
\end{equation}
\end{small}
Since we map time series of each entity into unique target distributions, different ranges of $S_{ck}$ are observed. This bias will assign each entity to different weights in terms of its contribution to $S_c$. To circumvent the above-unexpected bias, we design the entity-specific threshold for each entity. 
Considering different scales of $S_{ck}$, IQR is used to set respective thresholds. Therefore, the threshold for $S_{ck}$ is given as: $
    Threshold_{k} = \lambda_k(Q_3^k + 1.5*(Q_3^k - Q_1^k)),
    \label{equ:entityanomalyscore}
$
where $Q_1^k$ and $Q_3^k$ are 25-$th$ and 75-$th$ percentile of $S_{ck}$ across all observations, respectively. And $\lambda_k $ is used to adjust $Threshold_{k}$ because normal observations in different entities also fluctuate with different scales. 

\section{Experiment}
\subsection{Experiment Setup}
\subsubsection{Dataset}
The commonly used public datasets for MTS anomaly detection in OCC are MSL (Mars Science
Laboratory rover)~\cite{hundman2018detecting}, SMD (Server Machine Dataset)~\cite{su2019robust}, PSM (Pooled Server Metrics)~\cite{abdulaal2021practical}, SWaT (Secure Water Treatment)~\cite{goh2016dataset} and WADI (Water Distribution)~\cite{ahmed2017wadi}. 
The sensor data in SWaT and WADI is from water Treatment with 51 and 123 entities. MSL, SMD, and PSM are from Mars rover with 55 features, server metrics with 38 features, and server nodes at eBay with 25 features, respectively.
Since only normal time series are provided in these datasets for training in OCC setting, we follow the dataset setting of GANF~\cite{dai2021graph} and split the original testing dataset by 60\% for training, 20\% for validation, and 20\% for testing in SWaT. For other datasets, the training split contains 60\% data, and the test split contains 40\% data. 

\begin{table*}
    \footnotesize
    \centering
    \begin{tabular}{ccccccccc}
    \toprule
            Dataset &DeepSVDD &ALOCC &DROCC &DeepSAD & USAD & DAGMM &GANF &MTGFlow \\
        \midrule
       \rule{0pt}{10pt} SWaT &66.8\textpm{2.0} &77.1\textpm{2.3} &72.6\textpm{3.8} &75.4\textpm{2.4} &78.8\textpm{1.0}&72.8
\textpm{3.0} &79.8\textpm{0.7} & \textbf{84.8\textpm{1.5}} \\
       \rule{0pt}{10pt} WADI  & 83.5\textpm{1.6} &83.3\textpm{1.8} &75.6\textpm{1.6}& 85.4\textpm{2.7}  &86.1\textpm{0.9} &77.2\textpm{0.9}
& 90.3\textpm{1.0} &  \textbf{91.9\textpm{1.1}} \\ 
       \rule{0pt}{10pt} PSM  & 67.5\textpm{1.4} &71.8\textpm{1.3} &74.3\textpm{2.0}& 73.2\textpm{3.3} &78.0\textpm{0.2}  &64.6 \textpm{2.6} 
& 81.8\textpm{1.5} &  \textbf{85.7\textpm{1.5}} \\
       \rule{0pt}{10pt} MSL  & 60.8\textpm{0.4} &60.3\textpm{0.9} &53.4\textpm{1.6}& 61.6\textpm{0.6} &57.0\textpm{0.1}  &56.5 \textpm{2.6} 
& 64.5\textpm{1.9} &  \textbf{67.2\textpm{1.7}} \\ 
       \rule{0pt}{10pt} SMD  & 75.5\textpm{15.5} &80.5\textpm{11.1}& 76.7\textpm{8.7}  &85.9 \textpm{11.1} &86.9\textpm{11.7} &78.0\textpm{9.2}
& 89.2\textpm{7.8} &  \textbf{91.3\textpm{7.6}} \\ 
       \bottomrule	
    \end{tabular}
        \caption{Anomaly detection performance of AUROC(\%) on five public datatsets.}
    \label{tab:my_label}
\end{table*}


\subsubsection{Implementation Details}
For all datasets, we set the window size as 60 and the stride size as 10. Adam optimizer with a learning rate 0.002 is utilized to update all parameters. One layer of LSTM is sufficient to extract time representations in our experiment.
One self-attention layer with 0.2 dropout ratio is adopted to learn the graph structure.
We use MAF as the normalizing flow model.
For SWaT, one flow block and 512 batch size are employed. 
For other datasets, we arrange two flow blocks for it and set the batch size as 256. $\lambda$ is set as 0.8 for thresholds of all entities. The epoch is 40 for all experiments, which are performed in PyTorch-1.7.1 with a single NVIDIA RTX 3090 24GB GPU\footnote{ Code is available at github.com/zqhang/MTGFLOW.}.
\subsubsection{Evaluation Metric}
As in previous works, MTGFlow aims to detect window-level anomalies, and labels are annotated as abnormal when there exists any anomalous time point. The performance is evaluated by the Area Under the Receiver Operating Characteristic curve (AUROC).

\subsubsection{Baselines}
SOTA methods include semi-supervised methods DeepSAD~\cite{ruff2019deep}, OCC methods DeepSVDD~\cite{ruff2018deep}, ALOCC~\cite{sabokrou2020deep}, DROCC~\cite{goyal2020drocc}, and USAD~\cite{audibert2020usad} and unsupervised methods DAGMM~\cite{zong2018deep} and GANF~\cite{dai2021graph}. 
\subsubsection{Performance}





\begin{figure}
    \centering
    \includegraphics[width=1\columnwidth]{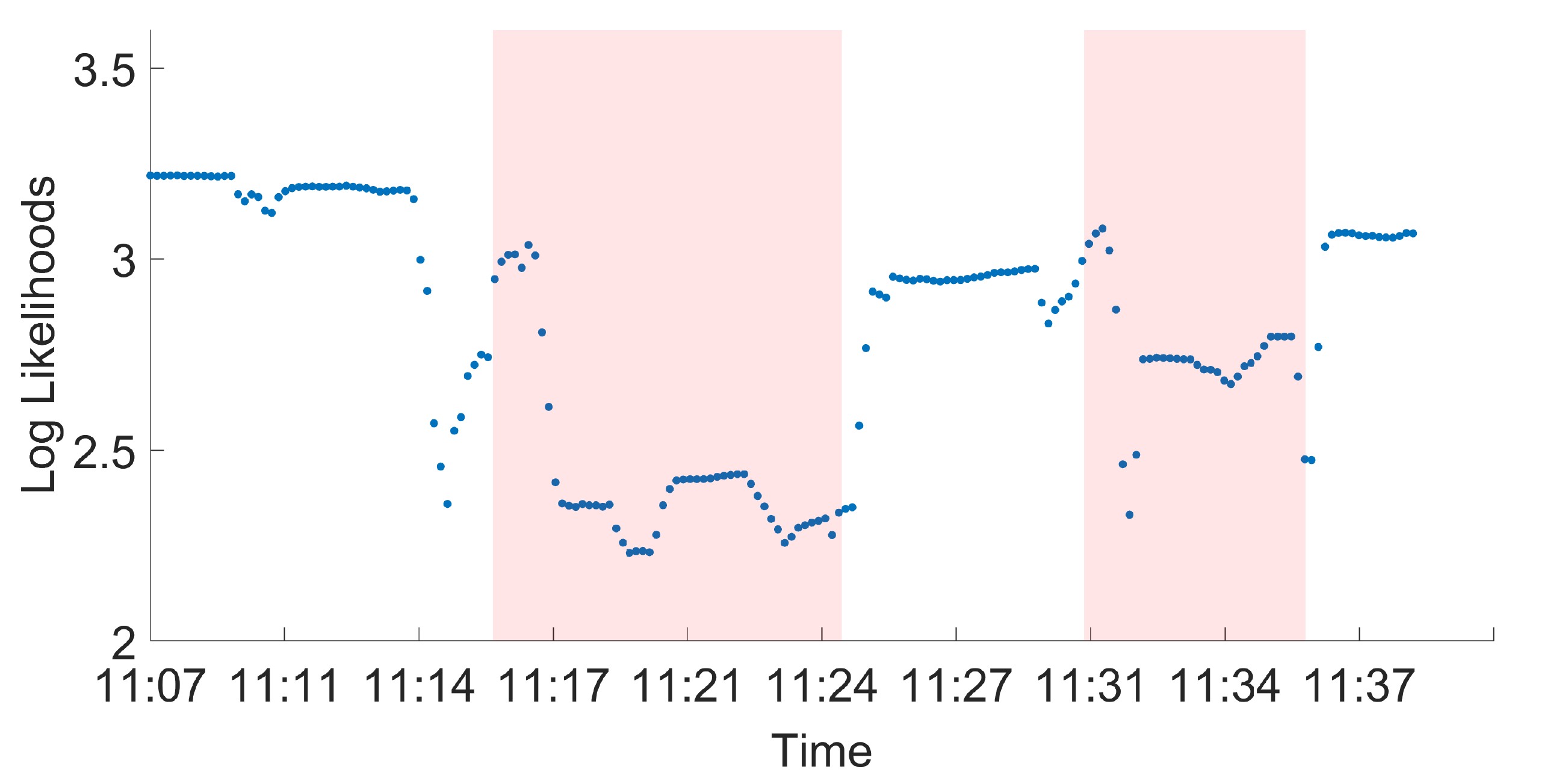}
    \caption{Log likelihoods for anomalies.}
    \label{fig: Log likelihoods for anomalies}
\end{figure}

\begin{figure}
    \centering
    \includegraphics[width=1\columnwidth]{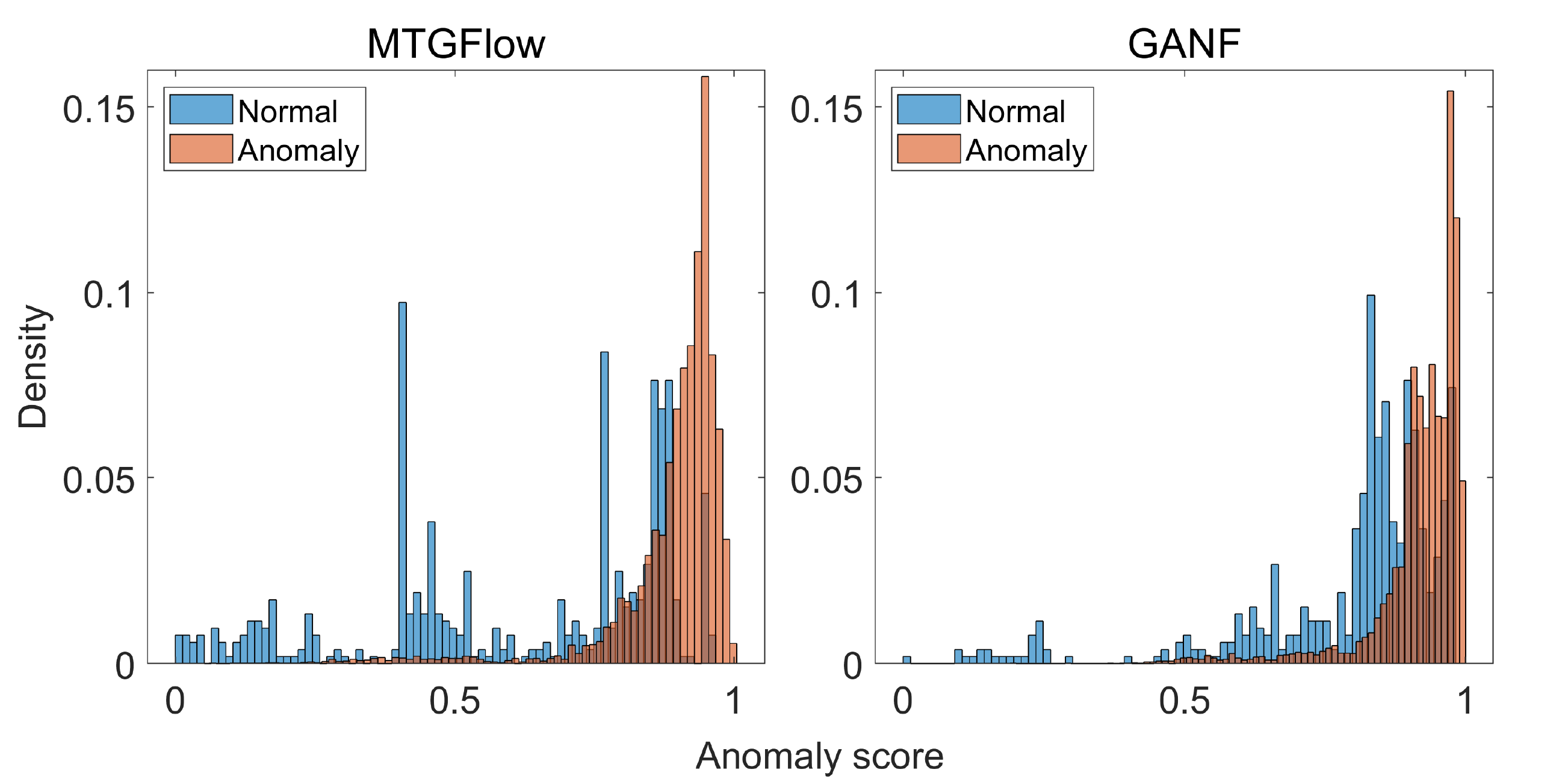}
    \caption{Comparison on normalized anomaly scores between MTGFlow and GANF.}
    \label{fig: comparison on anomaly scores between GANF and MTGFlow}
\end{figure}

We list the AUROC metric results in Table~\ref{tab:my_label}. Note that the standard deviation of SMD is large because it comprises 28 sub-datasets, where we test the performance on each of them and average all the results. MTGFlow has superior performance over all the other seven baselines. 
Compared with MTGFlow, DeepSVDD and DROCC project all training samples into the hypersphere so that they cannot learn the accurate decision boundary distinguishing normal from abnormal samples. Adversarial learning used by ALOCC and USAD and semi-supervised learning strategy in DeepSAD leverage a more informative training procedure to mitigate the effect of high anomaly contamination. As for DAGMM, it is restricted to the distribution estimation ability of GMM for multiple entities. Although GANF obtains a better result, its detection performance is still limited by inadequate dependence modeling and indiscriminative density estimation. 
Due to a much more flexible modeling structure, MTGFlow outperforms the above baseline methods.
Moreover, we study log likelihoods for anomalies ranging from 2016/1/2 11:07:00 to 11:37:00 in Fig.~\ref{fig: Log likelihoods for anomalies}. It is clear that log likelihoods are high for the normal series but lower for labeled abnormal ones (highlighted in red). This variation of log likelihoods validates that MTGFlow can detect anomalies according to low density regions of modeled distribution. Meanwhile, to investigate anomaly discrimination ability of MTGFlow, we present the normalized $S_c$ for MTGFlow and GANF in Fig.~\ref{fig: comparison on anomaly scores between GANF and MTGFlow}. As it is displayed, for normal series, anomaly scores of MTGFlow are more centered at 0 than these of GANF, and the overlap areas of normal and abnormal scores are also smaller in MTGFlow, reducing the false positive ratio. This larger score discrepancy corroborates that MTGFlow has superior detection performance.

\begin{table}
    \centering
    \footnotesize
    \begin{tabular}{cccccc}
    \toprule
         &Graph & Entity &SWaT &WADI  \\
        \midrule
       MTGFlow/(G, E)   &\XSolidBrush &\XSolidBrush & 78.3\textpm{0.9} & 89.7\textpm{0.5} \\
       MTGFlow/G   &\XSolidBrush &\Checkmark & 82.4\textpm{1.0} & 91.3\textpm{0.4}  \\
        MTGFlow/E  &\Checkmark&\XSolidBrush  & 81.2\textpm{1.1} & 91.0\textpm{0.7} \\
        MTGFlow     &\Checkmark& \Checkmark& \textbf{84.8\textpm{1.5}} & \textbf{91.9\textpm{1.1}}  \\
    \bottomrule
    \end{tabular}
        \caption{Module ablation study (AUROC\%).}
    \label{tab:ablation_module design}
\end{table}

\begin{table}
    \footnotesize
    \centering

    \begin{tabular}{ccccc}
    \toprule
        \multicolumn{2}{c}{\diagbox[width=10em]{Window size}{Blocks}}&1&2&3  \\\midrule
        \multirow{5}{*}{SWaT}&40&81.4\textpm{3.2}&82.7\textpm{2.1} &81.7\textpm{0.9}\\\cline{2-5}
        &60&\textbf{84.8\textpm{1.5}} &83.6\textpm{2.0}&83.1\textpm{0.9}\\\cline{2-5}
        &80&82.8\textpm{1.0}&82.7\textpm{0.8}&83.4\textpm{0.6}\\\cline{2-5}
        &100&82.6\textpm{0.5}&83.4\textpm{0.9}&83.5\textpm{0.6}\\\cline{2-5}
        &120&83.2\textpm{2.0} &83.4\textpm{2.3}&84.5\textpm{2.6}\\\cline{1-5}
        \multirow{5}{*}{WADI}&40&90.8\textpm{1.3}&91.7\textpm{1.2} &91.7\textpm{1.3}\\\cline{2-5}
        &60&89.2\textpm{1.9} &\textbf{91.9\textpm{1.1}}&91.5\textpm{0.8}\\\cline{2-5}
        &80&89.8\textpm{2.0}&90.7\textpm{0.8}&91.7\textpm{0.7}\\\cline{2-5}
        &100&89.6\textpm{1.1}&90.9\textpm{0.8}&91.8\textpm{0.6}\\\cline{2-5}
        &120&88.6\textpm{1.4} &91.0\textpm{0.6}&91.5\textpm{0.9}\\
     \bottomrule
    \end{tabular}
        \caption{Ablation study of hyperparameters (AUROC\%).}
    \label{tab:ablation_hyperparameters}
\end{table}

\subsection{Ablation Study}
\subsubsection{Module Ablation Study}
To test the validity of each designed module, we give several ablation experiments.
We denote MTGFlow without graph and entity-aware normalizing flow as MTGFlow/(G, E), MTGFlow only without graph as MTGFlow/G, and MTGFlow only without entity-aware normalizing flow as MTGFlow/E. Results are presented in Table~\ref{tab:ablation_module design}, where MTGFlow/(G, E) obtains the worst performance. It is attributed to the lack of relational modeling among entities and
indistinguishable density estimation. Applying graph structure learning to model pairwise relations, MTGFlow/E achieves better performance. Also, considering more fine-grained density estimation, MTGFlow/G achieves an improvement over MTGFlow/(G, E). Integrating these two modules, MTGFlow accomplishes the best results. 

\begin{figure}[t]
    \centering
    \includegraphics[width=1\columnwidth]{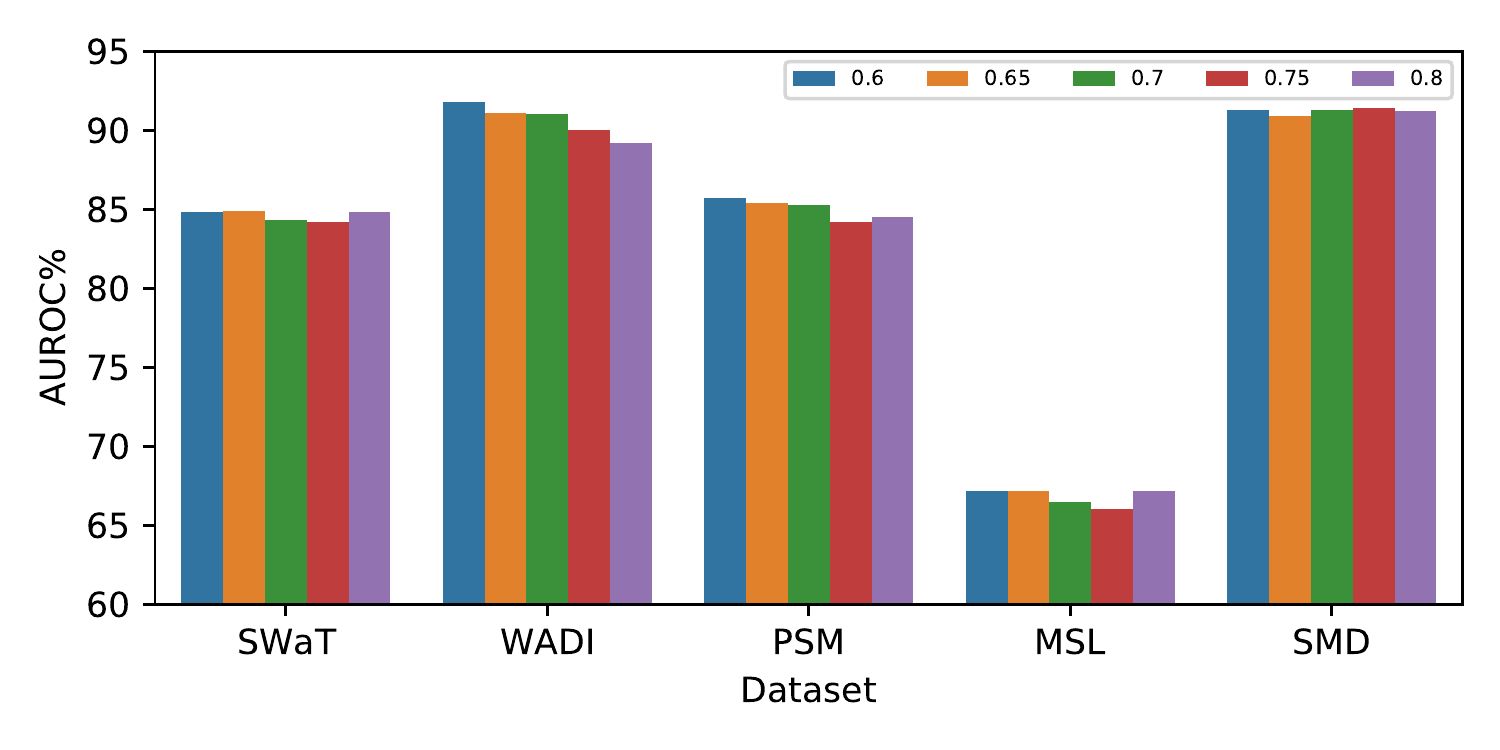}
    \caption{Effect of anomaly contamination ratio.}
    \label{fig: anomaly_ratio}
\end{figure}

\begin{figure*}
    \centering
    \includegraphics[width=1\textwidth]{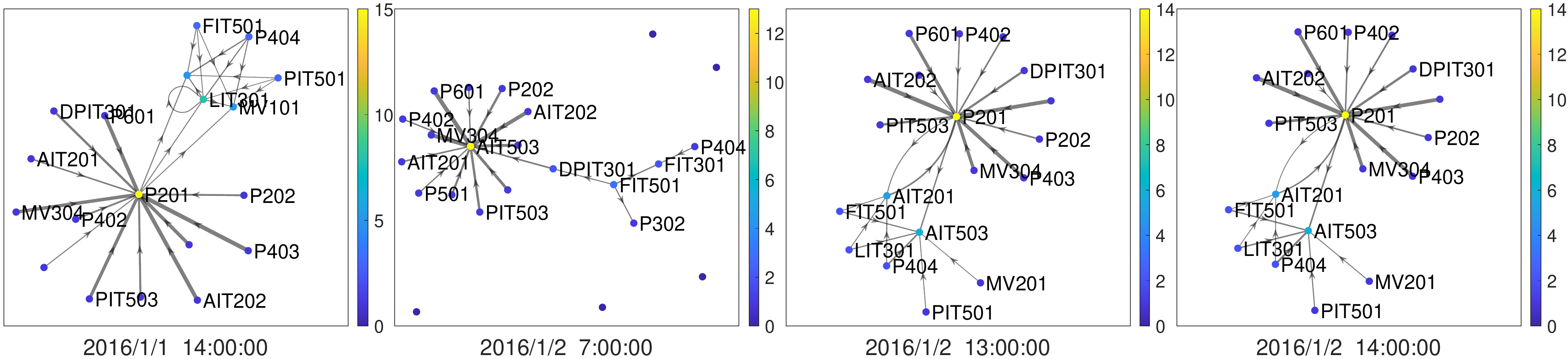}
    \caption{Dynamic graph structure in MTGFlow.}
    \label{fig:dynamic graph structure}
\end{figure*}


\subsubsection{Hyperparameter Robustness}
We conduct a comprehensive study on the choice of hyperparameters, the results are shown in Table~\ref{tab:ablation_hyperparameters}.
    Concretely, we conduct experiments with various sizes for the sliding window and the number of the normalizing flow blocks in Table~\ref{tab:ablation_hyperparameters}. When the window size is small, such as 40, 60, and 80, the increase in the number of blocks does not necessarily improve anomaly detection performance. A larger model may cause overfitting to the whole distribution, where abnormal sequences are undesirably located in high-density regions of this distribution. When the window size is larger, the distribution to be estimated is high-dimensional so that model needs more capacity. Hence, detection performance derives the average gain with blocks increasing due to more accurate distribution modeling. 


\subsubsection{Anomaly Ratio Analysis}
To further investigate the influence of anomaly contamination rates, we vary training splits to adjust anomalous contamination rates. 
For all the above-mentioned datasets, the training split increases from 60\% to 80\% with 5\% stride.
We present an average result over five runs in Fig.~\ref{fig: anomaly_ratio}. Although the anomaly contamination ratio of training dataset rises, the anomaly detection performance of MTGFlow remains at a stable high level.

\subsection{Result Analysis}
In order to further investigate the effectiveness of MTGFlow, we give a detailed analysis based on SWaT dataset.

\begin{figure}[t]
    \centering
    \includegraphics[width=1\columnwidth]{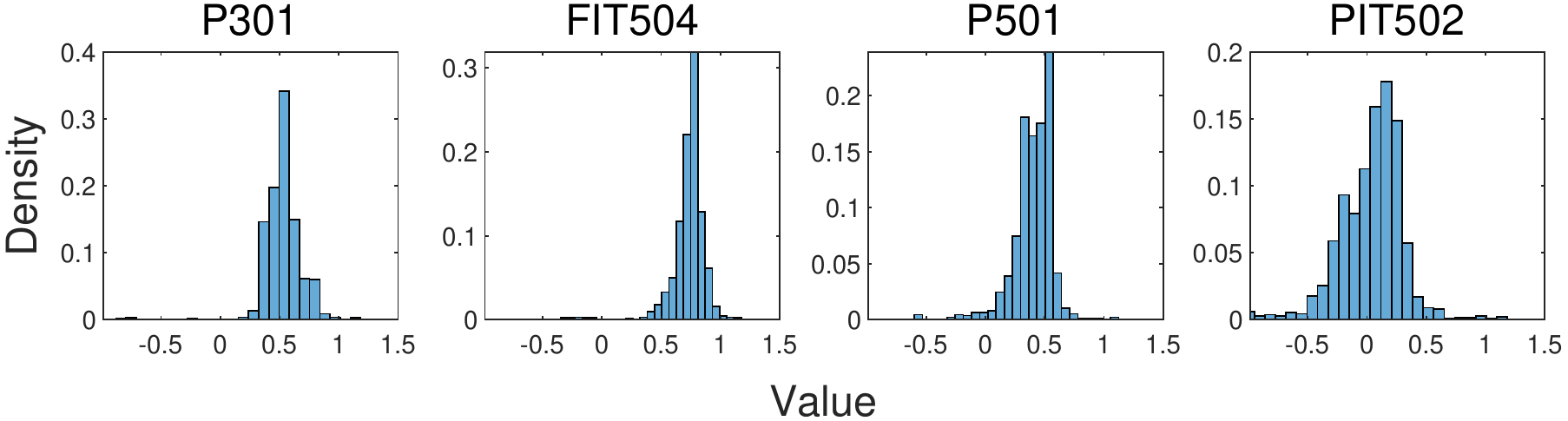}
    \caption{Transformed distributions of multiple entities.}
    \label{fig: Transformed distribution}
\end{figure}

\subsubsection{Dynamic Graph Structure}
Interdependencies among entities are not guaranteed to be immutable.
In fact, pairwise relations evolve with time. Benefiting from self-attention, MTGFlow can model this characteristic into a dynamic graph structure. We treat the attention matrix as the graph adjacent matrix. An empirical threshold of 0.15 is set for the adjacency matrix to show an intuitive learned graph structure in the test split. In Fig.~\ref{fig:dynamic graph structure}, the node size represents its node degrees, 
the arrow direction represents the learned directed dependence and the arrow width indicates the weight of the corresponding interdependencies.
The graph structure at 2016/1/1 14:00:00 is centered on the sensor $P201$, while the edges in the graph have completely changed and the center has shifted from $P201$ to $AIT503$ at 2016/1/2 7:00:00. This alteration of the graph structure may result from changing in working condition of water treatment plant.
 Besides, two similar graph structures can be found at 2016/1/2 13:00:00 and 2016/1/2 14:00:00. This suggests that the graph structure will be consistent if the interdependencies remain unchanged over a period of time, possibly due to repetitive work patterns of entities.
In addition, the main pairwise relations (thick arrow) at 2016/1/1 14:00:00 are similar as the ones at 2016/1/2 14:00:00, both centered on $P201$. It indicates that the interdependencies on multiple sensors are periodic. We also find mutual interdependencies from learned graph structures, such as the edges between $P201$ and $AIT201$ at 2016/1/2 13:00:00.
We summarize the findings: 
\begin{enumerate*}[label=(\arabic*)]
\item{Dynamic interdependencies among multiple entities.} 
\item{Consistent interdependencies among multiple entities.}  
\item{Periodic interdependencies among multiple entities.} 
\item{Mutual interdependencies among multiple entities.}
\end{enumerate*}
 Therefore, it is necessary to use a dynamic graph to model such changeable interdependencies.

\begin{figure}[t]
  \centering
    \includegraphics[width=1\columnwidth]{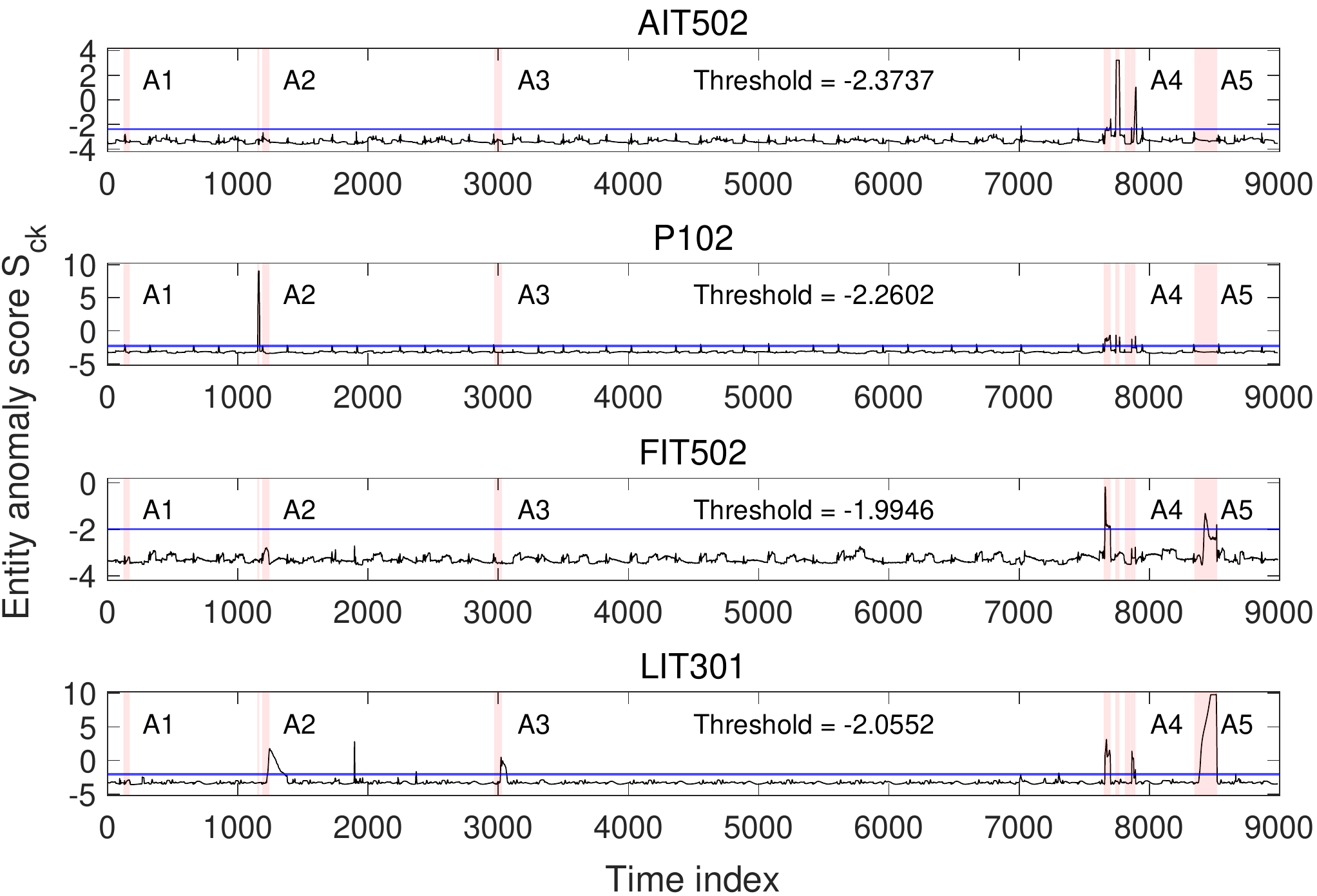}%
   
 \caption{Variation of log likelihoods for different entities on the whole testing dataset (anomalies are highlighted in red, and the blue line is the threshold according to $S_{ck}$).}
  \label{fig:sparse_characteristics}
\end{figure}

\subsubsection{Entity-specific Density Estimation} 
We further explore whether the distributions of all entities are transformed into different target distributions to verify our entity-aware design. 
Since the window size is 60, the corresponding transformed distributions are also 60-dimensional distributions.
Every single dimension of the multivariate Gaussian distribution is a Gaussian distribution. For better visualization, we present the 0-$th$ dimension of the transformed distributions in Fig.~\ref{fig: Transformed distribution}.
Four distributions of different entities are displayed. It can be seen that these distributions have been projected as unique distributions. Moreover, these distributions are successfully converted to preset Gaussian distributions with different mean vectors. 
The one-to-one mapping models entity-specific distributions and captures their respective sparse characteristics of anomalies. 



\subsubsection{Distinct Sparse Characteristics}

To demonstrate that the sparse characteristics vary with different entities, we study changes of $S_{ck}$ along time on SWaT. As shown in Fig.~\ref{fig:sparse_characteristics}, $S_{ck}$ of $AIT502$, $P102$, $FIT502$, and $LIT301$ are presented. The highlighted regions denote marked anomalies. For a better illustration, we divide the anomalous regions as $A_1$, $A_2$, $A_3$, $A_4$, and $A_5$ along the timeline. We can observe that different entities react to different anomalies because of their different work mechanisms. Specifically, $AIT502$ has obvious fluctuations at $A_4$, while $P102$ reacts to $A_2$. In addition, $FIT502$ is sensitive to $A_4$ and $A_5$, yet $LIT301$ is sensitive to $A_2$, $A_3$, $A_4$, and $A_5$. Nevertheless, MTGFlow is still able to accurately distinguish and detect these anomalies.
\section{Conclusion}
In this work, we proposed MTGFlow, an unsupervised anomaly detection approach for MTS based on the dataset with absolute zero known label. Extensive experiments on real-world datasets demonstrate its superiority, even if there exists high anomaly contamination. The superior anomaly detection performance of MTGFlow is attributed to dynamic graph structure learning and entity-aware density estimation. In addition, we explore various interdependencies that exist between individual entities from the learned dynamic graph structure. And a detected anomaly can be understood and localized via entity anomaly scores. In the future, we plan to explore the performance of our method from more realistic scenarios~\cite{wu2021current}, and further improve its practicality.

\section{Acknowledgements}
This work was supported by
the National Natural Science Foundation Program of China under Grant
U1909207 and Grant U21B2029.

\bibliography{aaai23}

\end{document}